\documentclass[letterpaper, 10 pt, conference]{ieeeconf}  

\IEEEoverridecommandlockouts                              

\usepackage{cite}
\usepackage{makecell}

\usepackage{amsmath,amssymb,amsfonts}
\usepackage{algorithmic}
\usepackage{graphicx}
\usepackage{textcomp}
\usepackage{booktabs}
\usepackage{xcolor}
\usepackage{dblfloatfix}
\usepackage{float}
\usepackage{threeparttable}

\usepackage{graphics} 
\usepackage{epsfig} 
\usepackage{mathptmx} 
\usepackage{times} 
\usepackage{amsmath}
\usepackage{amssymb}  

\begin{document}

\title{\LARGE \bf Traversability-Aware Cooperative Path Planning for Human–UGV Casualty Evacuation}

\author{Kristian Dalland $^{1}$, Prithvi Poddar $^{1}$, Souma Chowdhury $^{1,2}$, Karthik Dantu $^{2}$, Ehsan T. Esfahani $^{1}$
\thanks{This work is supported by the Office of Naval Research (award N00014-24-1-2003)}
\thanks{$^{1}$Department of Mechanical and Aerospace Engineering,
        University at Buffalo, NY 14260, USA, $^{2}$Department of Computer Science and Engineering,
        University at Buffalo, NY 14260, USA}
\thanks{Corresponding author: Ehsan T.~Esfahani (\texttt{ehsanesf@buffalo.edu})}
}

\maketitle
\thispagestyle{empty}
\pagestyle{empty}

\begin{abstract}

Heterogeneous multi-robot path planning is a well-studied problem in which agents with disparate kinematic and dynamic models must coordinate to achieve shared objectives. These formulations, however, treat all agents as robotic-their cost models are mechanical and their traversability is sensor-derived. In human-robot teaming, the human partner remains relegated to command and supervisory roles rather than being modeled as a physical co-navigator with distinct mobility constraints and dynamic energy reserves. This work investigates joint path planning for a two-agent human-UGV team in search-and-rescue casualty retrieval scenarios. We model the human agent using the Pandolf-Santee metabolic cost model with fatigue-modulated speed, and the UGV using a rolling-resistance energy model with terrain-dependent speed limits. By exploiting the complementary traversability of each agent-the human's ability to traverse dense vegetation and shallow water versus the UGV's superior speed on open terrain and roads-we optimize casualty transfer locations, termed switch points, to minimize total mission time. Evaluated across multiple synthetic $1\,\text{km}^2$ environments with procedurally generated elevation and land-cover data, the optimized strategy reduces mean mission time by 5.3\% relative to a human-only baseline and by 7.0\% relative to a naive human-UGV strategy without switch point optimization, while reducing human energy expenditure by 17.8\% relative to baseline. Notably, the naive strategy reduces human energy expenditure by a larger margin (22.4\%) but incurs a 2\% increase in mission time relative to baseline, illustrating that switch point optimization is necessary to realize time savings from human-UGV teaming.
\end{abstract}

\section{INTRODUCTION}
Planning for human--robot teams (HRT) is inherently challenging because the diversity of applications and robotic platforms, coupled with the unpredictability and latent heuristic nature of human behavior, introduces significant uncertainty into every stage of mission design \cite{goodrich2007hri}.
This difficulty is compounded by the dynamic, unstructured environments in which many HRT systems
must operate. Manufacturing is among the most visible domains: errors in
shared workspaces can result in injury or death for human operators, motivating
efforts to separate humans from robots entirely. Yet humans and robots bring 
complementary strengths to a task: robotic endurance, speed, and sensing paired 
with human reasoning, adaptability, and dexterity. Exploiting both yields gains 
in efficiency and capability that neither agent achieves alone. This 
complementary nature extends well beyond the factory floor to logistics,
search and rescue, autonomous driving, and disaster response
\cite{singh2026human, rizk2019task}.

Across these domains, HRT research has concentrated on task allocation,
communication, trust calibration, and the spectrum from shared to full
autonomy \cite{goodrich2007hri, sheridan2016hri}. The DARPA Subterranean
Challenge demonstrated that directed autonomy-where human operators set
waypoints and priorities while robots execute low-level control-outperformed
full autonomy across 16 real-world search missions, yielding more artifacts
found, greater distance traversed, and broader area coverage
\cite{robinson2024subt}. In human-centered tasks, integrating a
robotic partner improves mission outcomes by combining robotic perception and
endurance with human reasoning and decision-making, ultimately enhancing
performance while reducing risk to human operators \cite{adams2024human}.

Search-and-rescue robotics offers a particularly compelling case for
human-robot teaming. Early field deployments at the World Trade Center
demonstrated that robots served primarily as information sources rather than
autonomous actors, with effective coordination depending on shared mental
models between operators and robotic platforms \cite{casper2003hrt_wtc}.
Subsequent work by Murphy and Burke showed that human-robot teams with
higher situation awareness were up to nine times more likely to locate victims
than teams with lower awareness, highlighting that cognitive factors-not
robotic autonomy alone-are the primary bottleneck in field performance
\cite{murphy2005up}. More recent work on human-swarm interaction has
reinforced the centrality of the human element, showing that operator
physiological state directly affects coordination performance in
cyber-physical systems \cite{distefano2024exploring}. These findings
collectively establish that human-robot interaction in SAR is not merely a
matter of controlling robotic systems but of enabling teams to exploit robots
as active information sources, with communication, coordination, and operator
state playing critical roles.

Despite these advances, a notable gap persists. Heterogeneous multi-robot
path planning is well studied: ground-aerial teams
\cite{rizk2019task}, robots with differing kinematic constraints
\cite{solis2024adaptive}, and multi-platform coordination have all received significant attention. In these formulations, however, all 
agents are robotic-their cost models are mechanical and their 
traversability is sensor-derived, with human contribution limited to 
coordination and oversight. When a human appears in the loop, it 
is almost exclusively as an \emph{operator} directing robots 
\cite{distefano2024exploring}, not as a co-navigating \emph{agent} 
whose path must also be planned. The human's role is cognitive 
(task allocation, waypoint selection, supervisory control) rather 
than physical (carrying, modulating effort). Joint path planning that treats both the human and robot as physically traversing agents, with distinct mobility constraints and cost models, remains comparatively underexplored.

This work makes three primary contributions. First, we formulate joint path planning for a human–UGV team in which the human is explicitly modeled as a physical traversing agent with fatigue-modulated mobility. Second, we introduce a coordination framework based on optimizing intermediate transfer locations (switch points) under heterogeneous and non-comparable cost structures. Third, we demonstrate that heterogeneous teaming does not universally improve performance, and that coordinated planning is necessary to realize time savings in terrain-constrained environments.
\section{METHODS}
 
\subsection{System Overview}
 
This work explores joint planning for human and robotic agents in challenging terrain. We focus on path planning combined with casualty transfer between agent types over varying terrain types and gradients. Our agents consist of a two-person human team able to jointly carry a 100\,kg casualty in conjunction with a large off-/on-road UGV, modeled as the Clearpath Robotics Warthog \cite{clearpath_warthog}. These two agent types must navigate a complex environment to reach a 100\,kg casualty and return it to the starting location as quickly as possible. Additionally, we consider safety by restricting loaded paths to gentler terrain grades in scenarios where the casualty is severely injured.
 
\subsection{Agent Cost Models}
 
The human and UGV agents use fundamentally different traversal cost models. The human model is metabolic: it predicts physiological energy expenditure as a function of body mass, load, terrain, and grade, with speed modulated by cumulative fatigue. The UGV model is kinematic: it computes rolling resistance force from vehicle mass, slope, and surface type, with speed reduced by grade but unaffected by fatigue. Neither model produces costs in shared units, so joint planning optimizes over mission-level objectives-primarily total time-rather than summing edge costs across agents.
 
\subsubsection{Human Traversal Cost}
\label{sec:human_cost}
 
The human cost map for path planning is constructed using the Pandolf equation for energy expenditure \cite{pandolf1977}. This model combines body weight, carried load, terrain gradient, terrain type, and walking speed to produce a metabolic rate in watts:
\begin{equation}
    P = Mc_1 + (M + L)\left[c_2 \left(\frac{L}{M}\right)^2 + \eta\!\left(c_3 V^2 + c_4 VG\right)\right]
    \label{eq:pandolf}
\end{equation}
 
\begin{table}
\centering
\caption{Pandolf equation variables.}
\label{tab:PS_eq}
\begin{tabular}{llll}
\toprule
\textbf{Variable} & \textbf{Description} & \textbf{Unit} & \textbf{Type} \\
\midrule
$P$          & Metabolic rate        & W           & Output \\
$\eta$       & Terrain factor        & -         & Input \\
$M$          & Body mass             & kg          & Input \\
$L$          & External load         & kg          & Input \\
$V$          & Walking speed         & m/s         & Input \\
$G$          & Slope                 & \% grade    & Input \\
$c_1$-$c_4$ & Empirical constants   & -         & Fixed \\
\bottomrule
\end{tabular}
\end{table}
 
Once a candidate path has been generated, an accurate traversal time is computed using a human-energy-aware simulation \cite{dalland2025hybrid}. This hybrid system uses two models in sequence: first, the Pandolf equation with the Santee downhill correction \cite{santee2003load} calculates segment-wise energy expenditure; then, a three-tank hydraulic analogue of energy reserves \cite{weigend2023hydraulic} modulates walking speed based on cumulative fatigue. The hydraulic model abstracts energy reserves in a functional manner through a set of coupled ODEs that simulate exertion and recovery as a function of expenditure history. The result is a fatigue-aware speed profile in which the human slows progressively over long or strenuous segments and recovers partially during easier terrain.
 
\subsubsection{UGV Traversal Cost}
\label{sec:ugv_cost}
 
The UGV cost model computes two quantities per edge: traversal energy, used as the A* edge weight for path planning, and traversal time, accumulated along the planned path.
 
i) Edge Energy: For an edge of length $d$ (meters) with slope angle $\theta$, the resistive force is
\begin{equation}
    F = mg\!\left(\sin\theta + C_{rr}\cos\theta\right), \quad
    \theta = \arctan\!\left(\tfrac{g_\%}{100}\right),
    \label{eq:rolling_resistance}
\end{equation}
where $m$ is the vehicle mass, $g = 9.81\,\text{m\,s}^{-2}$, $g_\%$ is the percent grade, and $C_{rr}$ is the rolling resistance coefficient assigned by surface type (Table~\ref{tab:ugv_surface}). Edge cost is $E = \max(0,\, F \cdot d)$; the $\max(0,\cdot)$ term permits zero-cost coasting on steep descents.
 
Vehicle mass is differentiated by load state: $m = 280\,\text{kg}$ unloaded and $m = 380\,\text{kg}$ when carrying the 100\,kg casualty, consistent with the Clearpath Warthog specifications \cite{clearpath_warthog}.
 
ii) Edge Time: Traversal time is computed as a kinematic integral. The effective speed on each edge is
\begin{equation}
    v_{\text{eff}} = v_{\text{terrain}} \cdot \max\!\left(0.1,\; 1 - \frac{|g_\%|}{g_{\max}}\right),
    \label{eq:ugv_speed}
\end{equation}
where $v_{\text{terrain}}$ is the nominal speed for the edge's land-cover class (Table~\ref{tab:ugv_surface}) and $g_{\max}$ is the grade limit for the current state ($45\%$ unloaded, $20\%$ loaded). Speed decreases linearly with grade and is lower-bounded at 10\% of the nominal value. Total mission time is
\begin{equation}
    T = \sum_{i} \frac{d_i}{v_{\text{eff},i}}.
    \label{eq:ugv_time}
\end{equation}
 
This kinematic time model differs fundamentally from the human agent's traversal time, which emerges from a biophysical simulation of fatigue-modulated walking speed (Section~\ref{sec:human_cost}). The UGV experiences no fatigue: its speed depends only on terrain class and instantaneous grade. This asymmetry reflects the operational reality that human endurance degrades over distance while a battery-powered UGV maintains consistent output until energy is depleted. Mission duration is kept below the nominal runtime of 2.5 hours \cite{clearpath_warthog}.
 
iii) Relationship to Execution-Time Models: 
The rolling-resistance formulation is a planning-time simplification that operates on categorical land-cover data. Execution-time models such as the Stribeck friction-based cost function in AnyNav \cite{fu2025anynav} incorporate per-cell friction prediction from visual input, roughness, resistance, and turning penalties-inputs unavailable in a synthetic environment with procedurally generated terrain. In deployment, the categorical $C_{rr}$ and $v_{\text{terrain}}$ assignments would be replaced by continuously predicted friction and speed profiles from a visual perception model, enabling tighter coupling between planning and execution.
 
\subsubsection{Passability Constraints}
\label{sec:passability}
 
Hard impassability rules are enforced per agent type to reflect the distinct mobility limitations of human and UGV agents. Table~\ref{tab:passability} summarizes these constraints.
 
\begin{table}[H]
\centering
\caption{Agent-specific passability constraints.}
\label{tab:passability}
\begin{tabular}{lcc}
\toprule
\textbf{Constraint} & \textbf{Human} & \textbf{UGV (Warthog)} \\
\midrule
Water (depth $\leq 1.2\,\text{m}$)  & Passable   & Impassable \\
Water (depth $> 1.2\,\text{m}$)     & Impassable & Impassable \\
Dense forest                         & Passable   & Impassable \\
Slope, unloaded ($|g_\%| > 45\%$)    & Impassable & Impassable \\
Slope, loaded ($|g_\%| > 20\%$)      & Impassable & Impassable \\
\bottomrule
\end{tabular}
\end{table}
 
These differentiated constraints ensure that the UGV's traversable subgraph is a strict subset of the human's, which is the fundamental asymmetry motivating joint path planning: the human must traverse terrain the UGV cannot reach in order to retrieve the casualty and transport it to a feasible rendezvous point.
 
\subsection{Synthetic Environment}

\begin{figure*}[!t]
    \centering
    \includegraphics[width=\textwidth]{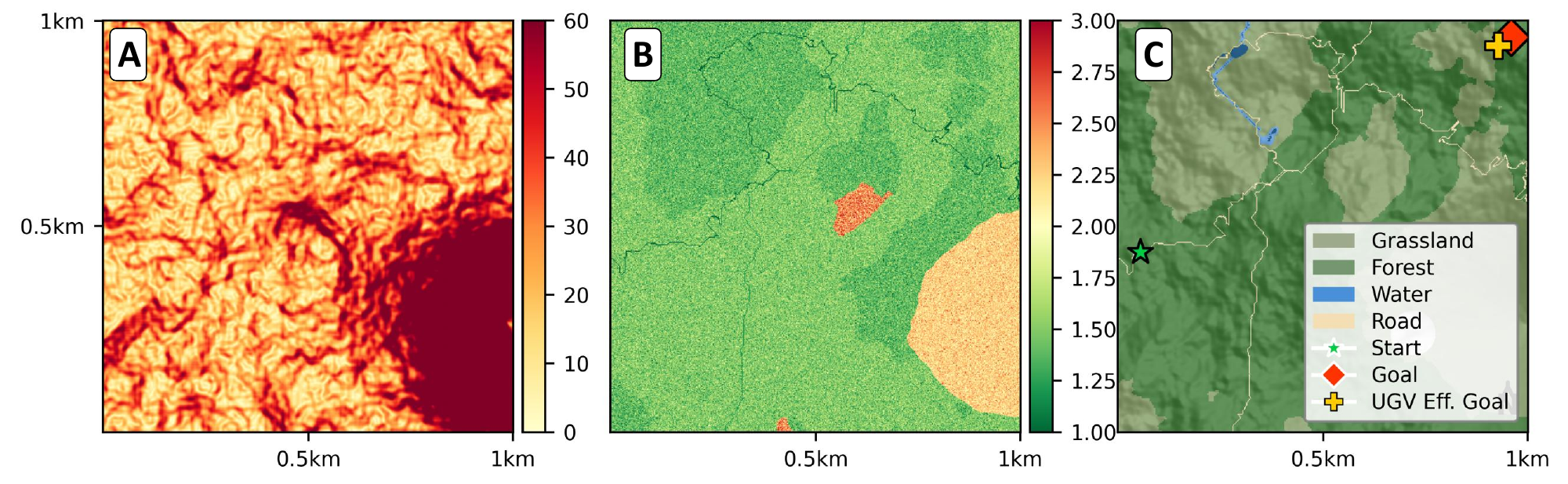}
    \vspace{-25 pt}
    \caption{Case 1. Synthetic environment overview. From left to right: (a)~terrain gradient map showing percent grade across the environment; (b)~human terrain factor ($\eta$) distribution derived from land-cover classification; (c)~composite land-cover map with start position, global goal, and effective UGV goal-the closest traversable cell to the global goal reachable by the Warthog platform.}
    \label{fig:synthetic_env}
\end{figure*}
 
In order to evaluate our planning framework, we designed a synthetic environment with realistic variance in terrain types and terrain grades. This decision was made to facilitate a high-resolution map, as publicly available digital elevation models and land-cover maps are limited to 30\,m resolution \cite{copernicus_dem_glo30, nalcms2020}. Our synthetic environment resolution is set to $2\,\text{m}$ per cell in each direction, which defines the per-pixel distance as well as the resolution of both the elevation model and land-cover classification. This resolution was also chosen to accommodate the Clearpath Warthog platform width of $1.38\,\text{m}$ \cite{clearpath_warthog}, ensuring that each agent always occupies a single cell.
 
\subsubsection{Grid Representation}
 
The environment is represented as a two-dimensional regular grid at a resolution of $2\,\text{m/cell}$. Each cell is connected to its eight neighbors via cardinal and diagonal adjacency. Edge attributes include Euclidean distance, computed as the cell spacing for cardinal edges and $\sqrt{2}$ times the cell spacing for diagonal edges, and slope grade (\%), derived from the elevation difference between adjacent cells.
 
\subsubsection{Elevation Model}
 
Terrain elevation is generated procedurally via fractional Brownian motion (fBm) using Perlin noise \cite{perlin1985} with six octaves, a persistence of $0.5$, and a lacunarity of $2.0$, following the terrain synthesis framework of Ebert et al.\ \cite{ebert2003}. The resulting noise field is scaled to a configurable elevation range to produce terrain profiles representative of hilly or mountainous environments. Slope grade at each cell is computed from central finite differences of the elevation surface.
 
\subsubsection{Land-Cover Generation}
\label{sec:landcover}
 
A second independent fBm noise layer \cite{perlin1985, ebert2003} partitions cells into discrete land-cover classes modeled after the 19-class NALCMS taxonomy \cite{nalcms2020}. These classes are collapsed into categories distinguishable by the agent cost models described above. Each class is assigned both a human terrain factor $\eta$ (Table~\ref{tab:terrain_factors}) and a UGV rolling resistance coefficient $C_{rr}$ with nominal speed $v_{\text{terrain}}$ (Table~\ref{tab:ugv_surface}).
 To introduce within-class heterogeneity, per-cell Gaussian noise ($\sigma = 0.05$) is added to each terrain factor assignment. Additionally, elevation-band gating constrains certain land-cover types to physically plausible locations: mud and bog are restricted to cells at or below the 25th elevation percentile.

\subsubsection{Road Network}
\label{sec:roads}

Search-and-rescue operations typically occur in environments with existing trail or road infrastructure; pure wilderness terrain without access corridors usually requires helicopter extraction rather than ground-based rescue. To reflect this, a configurable number of paved roads and dirt paths are generated procedurally between deterministic endpoints on opposing map edges, with the count and type of each specified as input parameters.

Each road is routed using A* with a cost function that combines Euclidean step distance with a soft grade penalty:

\begin{equation}
    c_{\text{road}} = d + \lambda \cdot \max\!\left(0,\; |g_\%| - g_{\text{road}}\right),
    \label{eq:road_cost}
\end{equation}

where $d$ is the edge distance in meters, $g_\%$ is the percent grade, $g_{\text{road}}$ is the maximum preferred grade for the road type ($8\%$ for paved, $15\%$ for dirt), and $\lambda = 800$ is a penalty weight. The heuristic is straight-line Euclidean distance to the goal. This formulation produces roads that follow low-grade corridors where possible and only exceed the preferred grade when no alternative route exists.
\begin{table}[t]
\centering
\caption{Terrain factor ($\eta$) assignments by land-cover class.}
\label{tab:terrain_factors}
\begin{threeparttable}
\begin{tabular*}{\linewidth}{@{\extracolsep{\fill}} lcl}
\toprule
\textbf{Land-Cover Class} & \textbf{$\eta$} & \textbf{Source} \\
\midrule
Paved road           & 1.00 & \cite{richmond2015}, Table 11 \\
Dirt / gravel road   & 1.20 & \cite{richmond2015}, Table 11 \\
Tundra               & 1.30 & \cite{richmond2015}, Table 11 \\
Grassland            & 1.40 & \cite{richmond2015}, Table 11 \\
Heavy brush / forest & 1.50 & \cite{soule1972} via \cite{richmond2015} \\
Mud / bog            & 2.50 & Interpolated\tnote{*} \\
\bottomrule
\end{tabular*}
\begin{tablenotes}
\small
\item[*] Midpoint interpolation between slippery clay ($\eta = 1.70$) and swamp ($\eta = 3.50$) from \cite{richmond2015}.
\end{tablenotes}
\end{threeparttable}
\end{table}
Cells along each resulting path are one cell wide and their land-cover class is overwritten with the corresponding road type, inheriting its terrain factor ($\eta = 1.0$ paved, $\eta = 1.2$ dirt) and rolling resistance coefficient ($C_{rr} = 0.015$ paved, $C_{rr} = 0.06$ dirt) from Tables~\ref{tab:terrain_factors} and~\ref{tab:ugv_surface}. Water crossings are handled as bridge edges with cost scaled by a bridge length multiplier, bypassing individual water cells.
 
i) Terrain Factors: Table~\ref{tab:terrain_factors} summarizes the terrain factor $\eta$ assigned to each land-cover class in human energy model.

ii) UGV Surface Coefficients: Table~\ref{tab:ugv_surface} summarizes the rolling resistance coefficient and nominal travel speed assigned to each land-cover class. The paved road $C_{rr}$ follows SAE~J2452 \cite{saej2452}; off-road values are engineering estimates within the ranges reported by Wong \cite{wong2008} for pneumatic tires on comparable surfaces, adjusted upward to reflect the wider tire footprint and lower inflation pressure of the Warthog platform. All forest classes, open water, and tundra/bog are marked impassable for the UGV, reflecting the platform's inability to traverse dense vegetation or deep water.
 
\begin{table}[h]
\centering
\caption{UGV surface coefficients by NALCMS land-cover class.}
\label{tab:ugv_surface}
\begin{threeparttable}
\begin{tabular}{lccc}
\toprule
\textbf{Surface Type} & \textbf{NALCMS} & \textbf{$C_{rr}$} & \textbf{$v_{\text{terrain}}$ (m/s)} \\
\midrule
Paved road             & -            & 0.015 & 4.0 \\
Dirt / gravel road     & -            & 0.06  & 2.5 \\
Wetland                & 14             & 0.10  & 1.5 \\
Grassland              & 10             & 0.12  & 1.8 \\
Cropland               & 15             & 0.18  & 1.2 \\
Shrubland              & 8              & 0.20  & 1.0 \\
Unclassified (default) & -            & 0.25  & 0.8 \\
\midrule
Forests, water, tundra/bog & 1,2,5,6,18,19 & \multicolumn{2}{c}{Impassable} \\
\bottomrule
\end{tabular}
\begin{tablenotes}
\small
\item $C_{rr}$ values are engineering estimates informed by Wong \cite{wong2008}. Nominal speeds reflect terrain difficulty and are reduced by grade as described in Section~\ref{sec:ugv_cost}.
\end{tablenotes}
\end{threeparttable}
\end{table}

\subsection{Path Planning}
Agent paths are generated using heap-based A* over the 8-connected grid. Two planning strategies are available, per loaded state and for each agent. The \textbf{Fast} strategy minimizes cumulative energy: metabolic cost via the Pandolf model \cite{pandolf1977} for the human and rolling resistance work (Eq.~\ref{eq:rolling_resistance}) for the UGV. The \textbf{Safe} strategy minimizes cumulative grade exposure by weighting each edge as $|g_\%| \times d$, producing conservative routes that favor gentle terrain at the expense of path length.

After A* returns a candidate path, a separate edge-by-edge simulation computes physically grounded traversal metrics. For the human, the full hybrid fatigue model (Section~\ref{sec:human_cost}) is applied iteratively along the path waypoints; travel speed is an output of this simulation rather than a fixed input. For the UGV, a kinematic walk accumulates $\sum d_i / v_{\text{eff},i}$ using the grade-penalized terrain speed (Eq.~\ref{eq:ugv_speed}). Both agents report total energy (kJ), travel time (s), path length (m), cumulative grade exposure, and maximum single-edge grade.

\subsection{Scenario Overview}
The evaluation scenario (Fig.~\ref{fig:synthetic_env}) defines a shared start position and a shared goal location for both agent types, and proceeds in two phases. In the \textbf{approach phase}, both agents depart from the shared start: the human team navigates to the goal location to retrieve the casualty, while the UGV travels to the first \textbf{switch point}, defined as a location where the casualty is handed off from one agent to the other.

In the \textbf{return phase}, the human team retrieves the 100\,kg casualty at the goal and must deliver it to the start as quickly as possible. The casualty is exchanged between agents at three switch points along the return route, with the human and UGV alternately bearing the load before the UGV delivers the casualty to the start position.

This return phase is the focus of joint planning optimization. The human is slower but can traverse dense forest and steep terrain that the UGV cannot. The UGV travels faster but is restricted to roads, paths, and open terrain such as grassland. This asymmetry creates opportunities to hand off the casualty at intermediate locations whose placement determines the efficiency of the team. The three switch points are optimized to minimize casualty evacuation time, with the human and UGV moving concurrently between handoffs and the earlier-arriving agent waiting for the other.

For comparison, the human-only baseline assigns the entire casualty return to the human team, while the naive human–UGV strategy transfers the casualty at the UGV-reachable location nearest the goal without optimizing switch-point locations.

\subsection{Switch Point Optimization}

Joint planning requires determining the switch-point locations that minimize total mission time.
Jointly deciding the number of switch points and their locations leads to an optimization problem with design variables of varying cardinality, and classical optimization solvers struggle with problems involving varying cardinality design variables. To mitigate this issue, we fix the number of switch points to $N=3$ in this work and optimize only their locations; jointly optimizing the number of switch points alongside their locations is left to future work.

For $N=3$ switch points, we formulate the problem of computing the switch point locations as a nonlinear optimization problem with the design variable defined as $\mathcal{P} = [\delta_1, \theta_1, \delta_2, \theta_2, \delta_3, \theta_3]$, where $\delta_i$ defines the distance (in meters) of switch point $i$ from switch point $i-1$, and $\theta_i$ defines the bearing (in degrees) of switch point $i$ from switch point $i-1$. The bearing angle $\theta_i$ is $0^\circ$ if the switch point $i$ lies on the straight line connecting the switch point $i-1$ to the start position. Any anti-clockwise deviation from this line leads to a positive bearing, and clockwise deviations result in negative bearings. The optimization is constrained such that none of the switch points lie in an area that is inaccessible by the UGV. For solving the optimization problem, we use the mixed discrete particle swarm optimization with diversity preservation, presented in \cite{MDPSO}. The particle swarm optimization is run for 100 iterations with a population size of 100 particles. We use the default algorithm parameter value as presented in \cite{MDPSO}.

\section{Results \& Discussion}

\begin{figure}
    \centering
    \includegraphics[width=1\linewidth]{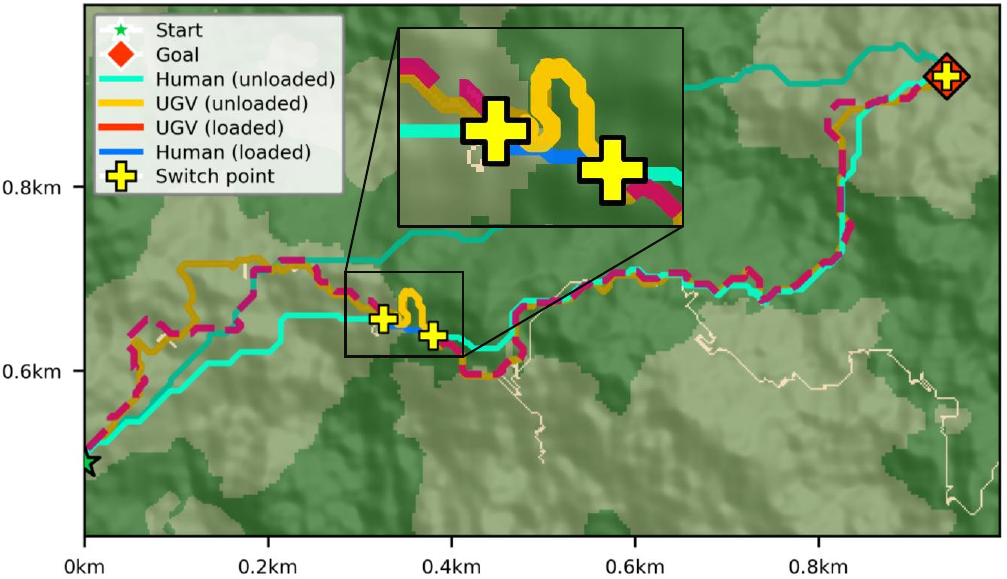}
    \vspace{-15pt}
    \caption{Optimized human-UGV collaborative routing for Case 2. The human
    carries the casualty through a short region impassable to the UGV
    while the UGV simultaneously routes around the obstacle, demonstrating how
    joint path optimization exploits complementary traversal capabilities to
    reduce total evacuation time.}
    \label{fig:seed43_collab}
\end{figure}

The core mechanism driving human-UGV collaboration is illustrated in Fig.~\ref{fig:seed43_collab}. In this representative scenario, the optimized strategy assigns the human to carry the casualty through a short region impassable to the UGV while the UGV simultaneously routes around the obstacle.
These switch points allow each agent to operate within its traversability envelope, and their concurrent execution is what produces mission time savings
over both the human-only and naive baselines.

\begin{figure}[t]
    \centering
    \includegraphics[width=\linewidth]{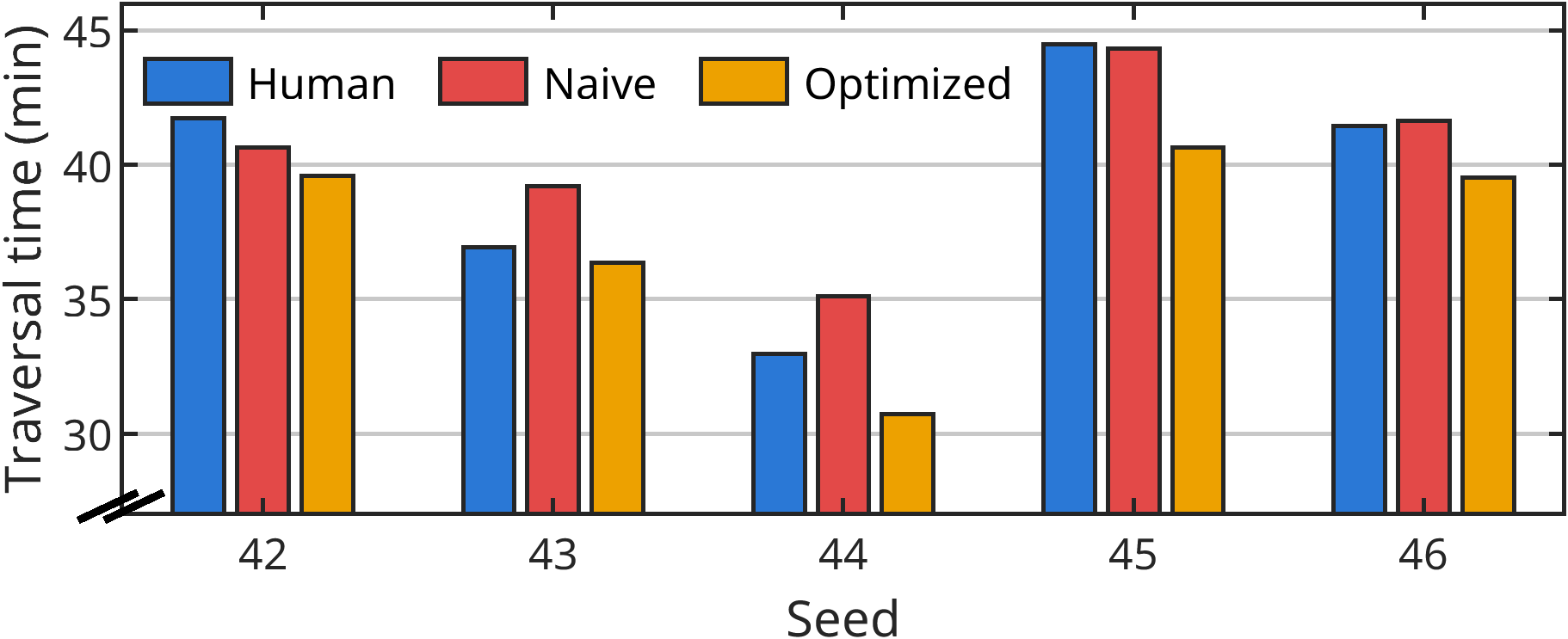}
    \vspace{-20pt}
    \caption{Total mission time per Case for the human-only baseline, naive
    human-UGV, and optimized human-UGV strategies. The optimized strategy
    achieves the shortest mission time in every scenario.}
    \label{fig:time_bar}
\end{figure}

Aggregate results across all Cases are summarized in Fig.~\ref{fig:time_bar}. The optimized strategy achieves the shortest mission time in every scenario, reducing mean mission time by 5.3\% relative to the human-only baseline and by 7.0\% relative to the naive human-UGV strategy. The naive strategy, by contrast, increases mean mission time by 2.0\% despite deploying a UGV teammate. This penalty is most pronounced in Cases 2 and 3 (Fig.~\ref{fig:time_bar}), where dense terrain forces the UGV to travel substantially longer distances than the direct human route, negating any speed advantage. The optimized strategy recovers time savings even in these constrained scenarios by solving for switch point locations that account for the asymmetric cost of each detour.

\begin{figure}[b]
    \centering
    \includegraphics[width=\linewidth]
    {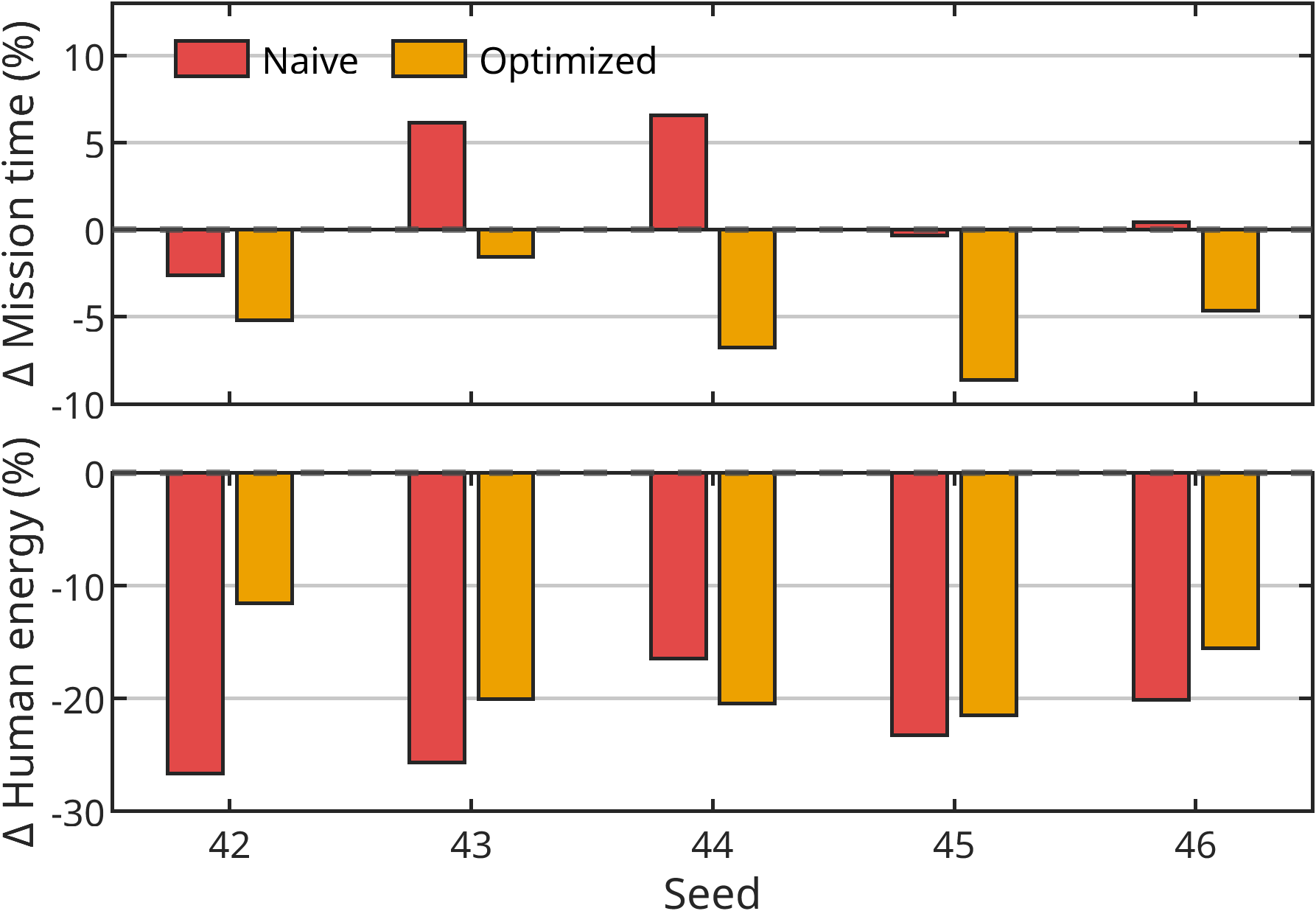}
    \caption{Percentage change relative to the human-only baseline for naive and optimized human–UGV strategies. (Top) Mission time: optimized reduces time by 5.3\% on average, while naive increases it by 2.0\%. (Bottom) Human energy: naive achieves larger reductions (22.4\%) than optimized (17.8\%), illustrating the time–energy trade-off.}
    \label{fig:percentages}
\end{figure}
The energy results in Fig.~\ref{fig:percentages} reveal an important trade-off.
Both strategies substantially reduce human energy expenditure relative to the
human-only baseline, but the naive strategy achieves a larger mean reduction
(22.4\%) than the optimized strategy (17.8\%). This is expected: the naive
strategy maximizes UGV loaded time regardless of routing cost, offloading the
most effort from the human but at the expense of a longer overall mission. The
optimized strategy accepts a slightly smaller reduction in human energy in
exchange for consistent time savings. In time-critical casualty evacuation,
where minutes can determine patient outcome, this trade-off favors optimization.
Case 3 is notable in that the optimized strategy outperforms the naive approach on both mission time and human energy expenditure simultaneously. By jointly optimizing all three switch points-including the initial human-UGV rendezvous-the framework effectively optimizes the UGV's goal location, allowing the planner to find a rendezvous that reduces the human carry distance and total mission time at once.

These results should be interpreted in the context of the scenario scale.
The $1\,\text{km}^2$ synthetic environments and short mission durations limit
the expression of two key cost drivers: human fatigue and UGV battery
consumption. In longer missions, fatigue-modulated speed degradation would
increasingly favor UGV casualty transfer, and energy constraints on both agents
would raise the stakes of suboptimal switch point placement. We expect the
performance gap between naive and optimized strategies to widen with mission
scale, making the optimization framework more consequential in operationally
realistic scenarios.

\section{Conclusion}

This work presents a joint path planning framework for human-UGV casualty retrieval that explicitly models the distinct traversability and energy dynamics of each agent. By optimizing switch point locations between the human and UGV, the framework consistently reduces total mission time across all evaluated scenarios, achieving a mean reduction of 5.3\% over a human-only baseline and 7.0\% over a naive human-UGV strategy. The results demonstrate that UGV teaming without switch point optimization can \textit{increase} mission time in terrain-constrained environments, and that optimization is necessary to reliably realize the benefits of heterogeneous teaming.

Beyond time savings, the framework also reduced human energy expenditure by 17.8\% on average. The trade-off between energy offloading and evacuation time shows that optimizing for speed and human effort are not equivalent, and the choice between them should reflect operational priorities.

Future work will extend the framework to larger environments where human fatigue and UGV battery life become binding constraints, and to multi-casualty scenarios with additional agents, where the number of switch points must itself be jointly optimized alongside their locations. Additional objectives, including injury-severity-weighted urgency, probabilistic terrain traversability, and physiological estimates of task difficulty are also planned \cite{distefano2021using}.

\addtolength{\textheight}{-12cm}   
\bibliographystyle{ieeetr}
\bibliography{ref}

@article{fu2025anynav,
  title={Anynav: Visual neuro-symbolic friction learning for off-road navigation},
  author={Fu, Taimeng and Zhan, Zitong and Zhao, Zhipeng and Du, Yi and Su, Shaoshu and Lin, Xiao and Esfahani, Ehsan Tarkesh and Dantu, Karthik and Chowdhury, Souma and Wang, Chen},
  journal={arXiv preprint arXiv:2501.12654},
  year={2025}
}

@inproceedings{dalland2025hybrid,
  title={Hybrid Human Model for Time of Hike Prediction},
  author={Dalland, Kristian and Wang, Chen and Dantu, Karthik and Chowdhury, Souma and Esfahani, Ehsan T},
  booktitle={2025 47th Annual International Conference of the IEEE Engineering in Medicine and Biology Society (EMBC)},
  pages={1--6},
  year={2025},
  organization={IEEE}
}

@misc{nalcms2020,
  author       = {{Commission for Environmental Cooperation}},
  title        = {North American Land Cover, 2020 ({Landsat}, 30m)},
  year         = {2020},
  howpublished = {North American Environmental Atlas},
  url          = {https://www.cec.org/north-american-environmental-atlas/land-cover-30m-2020/},
  note         = {Accessed: 2026-04-06}
}

@misc{copernicus_dem_glo30,
  author       = {{European Space Agency}},
  title        = {Copernicus Digital Elevation Model ({GLO}-30)},
  year         = {2024},
  howpublished = {Distributed by OpenTopography},
  doi          = {10.5270/ESA-c5d3d65},
  note         = {Accessed: 2026-04-06}
}

@misc{clearpath_warthog,
  author       = {{Clearpath Robotics}},
  title        = {Warthog Unmanned Ground Vehicle},
  year         = {2020},
  howpublished = {Product Data Sheet},
  url          = {https://clearpathrobotics.com/warthog-unmanned-ground-vehicle-robot/},
  note         = {Accessed: 2026-04-06}
}

@article{pandolf1977,
  author  = {K. B. Pandolf and B. Givoni and R. F. Goldman},
  title   = {Predicting energy expenditure with loads while standing or walking very slowly},
  journal = {Journal of Applied Physiology},
  volume  = {43},
  number  = {4},
  pages   = {577--581},
  year    = {1977}
}

@article{richmond2015,
  author  = {P. W. Richmond and A. W. Potter and W. R. Santee},
  title   = {Terrain factors for predicting walking and load carriage energy costs: Review and refinement},
  journal = {Journal of Sport and Human Performance},
  volume  = {3},
  number  = {3},
  year    = {2015}
}

@techreport{santee2003load,
  author      = {W. R. Santee and L. A. Blanchard and K. L. Speckman and J. A. Gonzalez and R. F. Wallace},
  title       = {Load carriage model development and testing with field data},
  institution = {US Army Research Institute of Environmental Medicine},
  address     = {Natick, MA},
  number      = {ADA 415788},
  year        = {2003}
}

@inproceedings{perlin1985,
  author    = {Perlin, Ken},
  title     = {An Image Synthesizer},
  booktitle = {Proc. 12th Annu. Conf. Comput. Graph. Interact. Tech. (SIGGRAPH)},
  pages     = {287--296},
  year      = {1985},
  doi       = {10.1145/325334.325247}
}

@book{ebert2003,
  author    = {Ebert, David S. and Musgrave, F. Kenton and Peachey, Darwyn and Perlin, Ken and Worley, Steven},
  title     = {Texturing and Modeling: A Procedural Approach},
  edition   = {3rd},
  publisher = {Morgan Kaufmann},
  year      = {2003},
  isbn      = {978-1558608481}
}

@article{soule1972,
  author  = {Soule, Ralph G. and Goldman, Ralph F.},
  title   = {Terrain Coefficients for Energy Cost Prediction},
  journal = {J. Appl. Physiol.},
  volume  = {32},
  number  = {5},
  pages   = {706--708},
  year    = {1972},
  doi     = {10.1152/jappl.1972.32.5.706}
}

@book{wong2008,
  author    = {Wong, J. Y.},
  title     = {Theory of Ground Vehicles},
  edition   = {4th},
  publisher = {John Wiley \& Sons},
  year      = {2008},
  isbn      = {978-0-470-17038-0}
}

@misc{saej2452,
  author       = {{SAE International}},
  title        = {Stepwise Coastdown Methodology for Measuring Tire Rolling Resistance},
  howpublished = {SAE Standard J2452},
  month        = jun,
  year         = {1999},
  note         = {Revised Jul.\ 2017},
  doi          = {10.4271/J2452_201707}
}

@article{weigend2023hydraulic,
  author  = {F. C. Weigend and D. C. Clarke and O. Obst and J. Siegler},
  title   = {A hydraulic model outperforms work-balance models for predicting recovery kinetics from intermittent exercise},
  journal = {Annals of Operations Research},
  volume  = {325},
  number  = {1},
  pages   = {589--613},
  year    = {2023}
}

@article{singh2026human,
  title={Human--Robot Teaming: A Comprehensive Survey on Collaboration, Communication, and Cognition},
  author={Singh, Saurav and Rantanen, Esa and Heard, Jamison},
  journal={ACM Transactions on Human-Robot Interaction},
  volume={15},
  number={2},
  pages={1--48},
  year={2026},
  publisher={ACM New York, NY}
}

@inproceedings{murphy2005up,
  title={Up from the rubble: Lessons learned about HRI from search and rescue},
  author={Murphy, Robin R and Burke, Jennifer L},
  booktitle={Proceedings of the Human Factors and Ergonomics Society Annual Meeting},
  volume={49},
  number={3},
  pages={437--441},
  year={2005},
  organization={SAGE Publications Sage CA: Los Angeles, CA}
}

@article{adams2024human,
  title={Human--robot teaming challenges for the military and first response},
  author={Adams, Julie A and Scholtz, Jean and Sciarretta, Albert},
  journal={Annual Review of Control, Robotics, and Autonomous Systems},
  volume={7},
  year={2024},
  publisher={Annual Reviews}
}

@article{solis2024adaptive,
  title={Adaptive robot coordination: A subproblem-based approach for hybrid multi-robot motion planning},
  author={Solis, Irving and Motes, James and Qin, Mike and Morales, Marco and Amato, Nancy M},
  journal={IEEE Robotics and Automation Letters},
  volume={9},
  number={8},
  pages={7238--7245},
  year={2024},
  publisher={IEEE}
}

@article{goodrich2007hri,
  author  = {Goodrich, Michael A. and Schultz, Alan C.},
  title   = {Human--Robot Interaction: A Survey},
  journal = {Found. Trends Hum.-Comput. Interact.},
  volume  = {1},
  number  = {3},
  pages   = {203--275},
  year    = {2007},
  doi     = {10.1561/1100000005}
}

@article{sheridan2016hri,
  author  = {Sheridan, Thomas B.},
  title   = {Human--Robot Interaction: Status and Challenges},
  journal = {Hum. Factors},
  volume  = {58},
  number  = {4},
  pages   = {525--532},
  year    = {2016},
  doi     = {10.1177/0018720816644364}
}

@article{rizk2019task,
  author  = {Rizk, Yara and Awad, Mariette and Tunstel, Edward W.},
  title   = {Cooperative Heterogeneous Multi-Robot Systems: A Survey},
  journal = {ACM Comput. Surv.},
  volume  = {52},
  number  = {2},
  pages   = {1--31},
  year    = {2019},
  doi     = {10.1145/3303848}
}

@article{distefano2024exploring,
  title={Exploring human-swarm interaction dynamics in cyber-physical systems: A physiological approach},
  author={Distefano, Joseph P and Chowdhury, Souma and Esfahani, Ehsan},
  journal={Journal of Integrated Design and Process Science},
  volume={27},
  number={3-4},
  pages={200--210},
  year={2024},
  publisher={SAGE Publications Sage UK: London, England}
}

@article{casper2003hrt_wtc,
  author  = {Casper, Jennifer and Murphy, Robin R.},
  title   = {Human--Robot Interactions During the Robot-Assisted Urban Search and Rescue Response at the {World Trade Center}},
  journal = {IEEE Trans. Syst., Man, Cybern. B},
  volume  = {33},
  number  = {3},
  pages   = {367--385},
  year    = {2003},
  doi     = {10.1109/TSMCB.2003.811794}
}

@article{robinson2024subt,
  author  = {Robinson, Nicole and Williams, Jason and Howard, David and Tidd, Brendan and Talbot, Fletcher},
  title   = {Human-Robot Team Performance Compared to Full Robot Autonomy in 16 Real-World Search and Rescue Missions: Adaptation of the {DARPA} Subterranean Challenge},
  journal = {ACM Trans. Hum.-Robot Interact.},
  volume  = {14},
  number  = {1},
  year    = {2024},
  doi     = {10.1145/3699955}
}

@Article{MDPSO,
author={Chowdhury, Souma
and Tong, Weiyang
and Messac, Achille
and Zhang, Jie},
title={A mixed-discrete Particle Swarm Optimization algorithm with explicit diversity-preservation},
journal={Structural and Multidisciplinary Optimization},
year={2013},
month={Mar},
day={01},
volume={47},
number={3},
pages={367-388},
issn={1615-1488},
doi={10.1007/s00158-012-0851-z},
url={https://doi.org/10.1007/s00158-012-0851-z}
}

@inproceedings{distefano2021using,
  title={Using physiological information to classify task difficulty in human-swarm interaction},
  author={Distefano, Joseph P and Manjunatha, Hemanth and Chowdhury, Souma and Dantu, Karthik and Doermann, David and Esfahani, Ehsan T},
  booktitle={2021 IEEE International Conference on Systems, Man, and Cybernetics (SMC)},
  pages={1198--1203},
  year={2021},
  organization={IEEE}
}

\end{document}